\documentclass[10pt,twocolumn,letterpaper]{article}

\usepackage{cvpr}
\usepackage{times}
\usepackage{epsfig}
\usepackage{graphicx}
\usepackage{amsmath}
\usepackage{amssymb}
\usepackage{booktabs}
\usepackage{color}

\usepackage[pagebackref=true,breaklinks=true,letterpaper=true,colorlinks,bookmarks=false]{hyperref}

\cvprfinalcopy 

\def\cvprPaperID{****} 

\ifcvprfinal\fi
\begin{document}

\title{Zoom-RNN: A Novel Method for Person Recognition Using Recurrent Neural Networks}

\author{Sina Mokhtarzadeh Azar\and Sajjad Azami\and Mina Ghadimi Atigh\and Mohammad Javadi\and Ahmad Nickabadi\\
CEIT Department, Amirkabir University of Technology\\
424 Hafez Ave, Tehran, Iran\\
{\tt\small \{sinamokhtarzadeh,sajadazami,minaghadimi,mohammad.javadi,nickabadi\}@aut.ac.ir}
}

\maketitle
\thispagestyle{plain}
\pagestyle{plain}

\begin{abstract}
The overwhelming popularity of social media has resulted in bulk amounts of personal photos being uploaded to the internet every day. Since these photos are taken in unconstrained settings, recognizing the identities of people among the photos remains a challenge.
Studies have indicated that utilizing evidence other than face appearance improves the performance of person recognition systems. In this work, we aim to take advantage of additional cues obtained from different body regions in a zooming in fashion for person recognition.
Hence, we present Zoom-RNN, a novel method based on recurrent neural networks for combining evidence extracted from the whole body, upper body, and head regions. Our model is evaluated on a challenging dataset, namely People In Photo Albums (PIPA), and we demonstrate that employing our system improves the performance of conventional fusion methods by a noticeable margin.
\end{abstract}

\section{Introduction}\label{introduction}

During the past decades, taking personal photos in daily life has become easier and more common with the advent of smartphones and digital cameras. Massive amounts of these personal images are uploaded to the internet, mostly through social media. Given that most of the times these images contain people, smart platforms are interested in the organization of identities in these photos. To perform the person recognition task, the question of "what is the identity of this person?" should be answered \cite{schroff2015facenet}.

\begin{figure*}
\centering
\includegraphics[height =8cm]{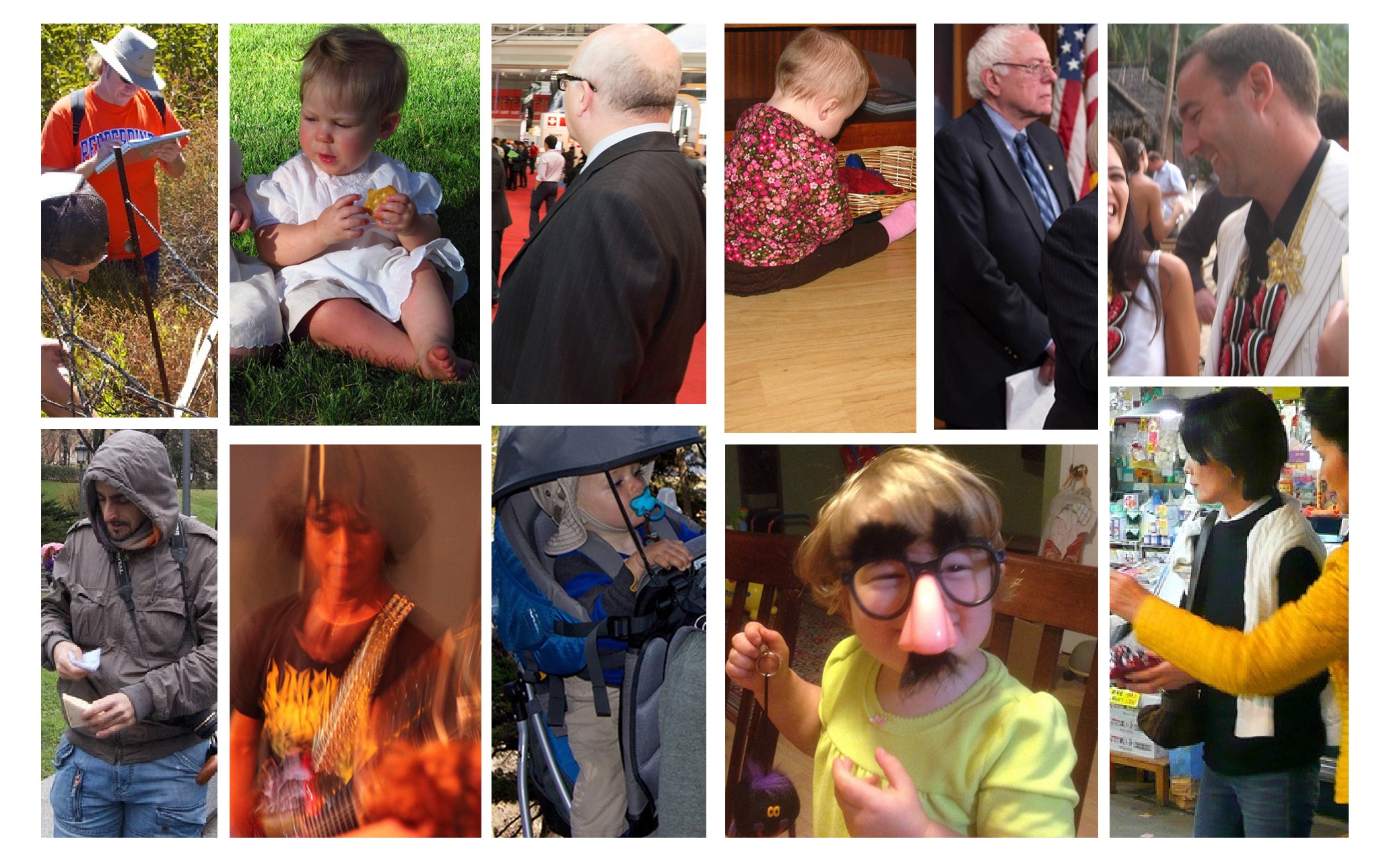}
\caption{Sample images of PIPA dataset. Occlusion, pose variance, low resolution and motion blur are challenging for the person recognition task}
\label{fig:ds}
\end{figure*}

The first person recognition models were developed based on hand-crafted features and were tested on constrained tiny datasets \cite{huang2007labeled,guillaumin2009you,zhao2013person}. But, these models cannot be easily applied to the problem of person recognition in photo album settings due to various challenges like occlusion, viewpoint changes, pose variance and low resolution represented by People in Photo Album (PIPA) \cite{zhang2015beyond}. Sample images of PIPA are shown in Figure \ref{fig:ds}.

There have been numerous studies like \cite{anguelov2007contextual,taigman2014deepface,schroff2015facenet,o2009context,lin2010joint} on person recognition in photo albums. The main ideas are to extract more sophisticated features from or about the input image and to employ more advanced classification methods for learning the relations between the features and identities.

Regarding the information sources used in the literature of person recognition task, some studies focus on relational nature of photos in an album belonging to an identity. Perhaps the most obvious way to capture this relation is to extract additional information from the photo. Contextual cues such as clothes, glasses, and surrounding objects, or even metadata like photo location and social relationship of identities, can drastically help the inference about an identity present in a photo album. Extraction, exploitation, and fusion of such information are extensively studied in previous works \cite{li2016multi,anguelov2007contextual,lin2010joint}.
Moreover, current person recognition methods, like many other image processing techniques, enjoy the informative representations of the input images provided by the convolutional neural networks (CNNs).

Human body parts, other than the face, are another source of information beneficial for identifying a person.

As discussed in studies like \cite{zhang2015beyond,oh2015person}, we have observed that relying on facial features in person recognition have shortcomings, specifically in dealing with non-frontal views or cluttered faces, which frequently happens in personal photos.

\textit{head}, \textit{upper body} and \textit{whole body} are the main body regions used in many person recognition models. However, the models differ in the way they aggregate the information extracted from these regions. In the \textit{early fusion} approach \cite{oh2015person}, the feature vectors extracted from different parts are combined to form the final descriptor used for the classification while in the \textit{late fusion}  \cite{li2016multi}, each feature vector is separately classified to form a probability vector on different identities and these initial decision vectors are then aggregated.

In this paper, we propose a novel fusion method, called zooming RNN, for combining the evidence extracted from main human body parts; head, upper body, and whole body. The proposed model incorporates both approaches of early-stage decision making based on the evidence obtained from each part and the late identification based on the final aggregated feature vector. To do so, a two-part recurrent neural network is applied to the feature and probability vectors extracted by convolutional neural networks from different regions of the human body. Experimental results on PIPA dataset show the superiority of the proposed model over other fusion mechanisms. The proposed model can be easily generalized to include more contextual information in recognition.

The rest of this paper proceeds as follows. After an overview of related works in Section \ref{related}, we describe and formulate our approach in Section \ref{approach}. In Section \ref{experiments}, the evaluation benchmark, implementation details, experimental procedures and results are presented and compared to other methods. We provide visualizations of our predictions in Section \ref{visualization}, and conclude our work in Section \ref{conclusion}.

\section{Related Works}\label{related}

Since this paper proposes a person recognition model evaluated on PIPA dataset, the previous works related to the proposed model are reviewed in the following three subsections: (1) person recognition in photo album, (2) the person recognition models on PIPA, and (3) dependency modeling with RNNs.

\subsection{Person Recognition in Photo Album}

Person recognition in photo album is the task of identifying people in daily life photos such as social media or private photo collections \cite{liu2017rethinking}.
Recognition in photo album setting includes challenges like cluttered background, pose variance, age gap, and diverse clothing \cite{oh2015person,liu2017rethinking}.
The success of traditional face recognition algorithms was limited when applied on personal photos that are usually taken under uncontrolled conditions with significant variations in pose, expression, and illumination \cite{anguelov2007contextual}.

Anguelov \textit{et al.} \cite{anguelov2007contextual} used additional cues present in photo collections such as clothing and album metadata to provide context, employing a Markov Random Field (MRF) with similarity potentials, and tested the system on a relatively small dataset.
O\textsc{\char13}Hare \textit{et al.} \cite{o2009context} conducted a comprehensive empirical study using the real private photo collections of a number of users and proposed language modeling and nearest neighbor approaches to context-based person identification.
Lin \textit{et al.} \cite{lin2010joint} presented a probabilistic framework in which the relations between different domains (people, events, and locations) are estimated based on the co-occurrence information of the instances of two domains. The tagged objects of two other domains are used as the context for identification of an unknown object in the third domain.

Recent advances in processing power\cite{dean2012large} alongside the immense availability of large labeled datasets, e.g. Labeled Faces in the Wild (LFW) dataset \cite{huang2007labeled} with various challenges due to pose invariance, motion blur, and deformation, resulted in a need for scaling up learning techniques. Recently, deep neural networks have shown great performance in many computer vision tasks including person recognition. Taigman \textit{et al.} \cite{taigman2014deepface} trained their model on a large dataset and achieved accuracies around 97.45\% on LFW. Schroff \textit{et al.} \cite{schroff2015facenet} employed a data-driven method based on learning a Euclidean embedding per image using a deep convolutional network and achieved 99.63\% on LFW dataset. Sun \textit{et al.} \cite{wst2008deeply} achieved new state-of-the-art results on LFW \cite{huang2007labeled} and YouTube Faces \cite{wolf2011face} benchmarks by designing DeepID2+, increasing the dimension of hidden representations and adding supervision to early convolutional layers.

\subsection{Person Recognition Models on PIPA}

Studies like \cite{taigman2014deepface} , \cite{schroff2015facenet} and \cite{wst2008deeply} resulted in significant error reduction and approached human-level performance on commonly used standard datasets such as LFW. Recently, Zhang \textit{et al.} \cite{zhang2015beyond} have introduced People In Photo Album (PIPA) as a novel dataset addressing the limitations of conventional person recognition systems, most of which lied heavily on facial cues. PIPA has become a popular benchmark for person recognition ever since and various studies \cite{zhang2015beyond,oh2015person,li2016multi,li2017sequential,kumar2017pose,liu2017rethinking} have been conducted to reduce error on this dataset, each focusing on certain challenges. Along with the original dataset, the baseline accuracies were provided using a novel method called PIPER which significantly outperformed DeepFace \cite{taigman2014deepface} and AlexNet \cite{krizhevsky2012imagenet} on PIPA. In order to better challenge the generalization across long-term appearance changes of a person, Oh \textit{et al.} \cite{oh2015person} extended PIPA dataset and proposed 3 new splits. They also achieved better results on PIPA by evaluating the effectiveness of different body regions, the scene context and some attributes like age and gender.

The method introduced in \cite{oh2015person} was extended in \cite{oh2016faceless} with a concern on privacy issues of social media, with results indicating that only a handful of images are enough to threaten users’ privacy, even in the presence of obfuscation. Li \textit{et al.} \cite{li2016multi} went beyond single photo and presented a framework that exploits contextual cues at personal, group and photo levels, aiming at improving the recognition rate. Kumar \textit{et al.} \cite{kumar2017pose} proposed a network that jointly optimizes a single loss over multiple body regions to tackle the pose variations challenge. Liu \textit{et al.} \cite{liu2017rethinking} proposed a congenerous cosine loss, which optimizes the cosine distance among data features to simultaneously enlarge inter-class variation and intra-class similarity. They carried out experiments on various large-scale benchmarks including PIPA \cite{zhang2015beyond} and demonstrated the effectiveness of their algorithm.

\subsection{Dependency Modeling with RNNs.}

Sequence modeling approaches in many contexts benefit from recurrent architectures, particularly LSTMs \cite{hochreiter1997long} and GRUs \cite{chung2014empirical} due to the ability of these networks in modeling dependencies within sequences \cite{li2017sequential}. Recurrent Neural Networks (RNNs) have been extensively used in tasks like image captioning \cite{mao2014deep,kiros2014unifying,vinyals2015show} and language modeling \cite{sutskever2014sequence}. For our application, we are interested in extracting the relation between different body region features and person identities using RNNs. The most accurate study of relational cues on PIPA dataset are conducted by Li \textit{et al.} \cite{li2017sequential}. They focus on relational information between people in the same photo, use the scene context and employ an RNN to achieve state-of-the-art results. Similarly, \cite{wang2016cnn} and \cite{stewart2016end} exploited the label dependencies in an image based on decoding an image into a set of people detections.

\section{Our Approach}\label{approach}
\begin{figure}
\centering
\includegraphics[trim = 9 475 0 8,clip,height = 5.5cm]{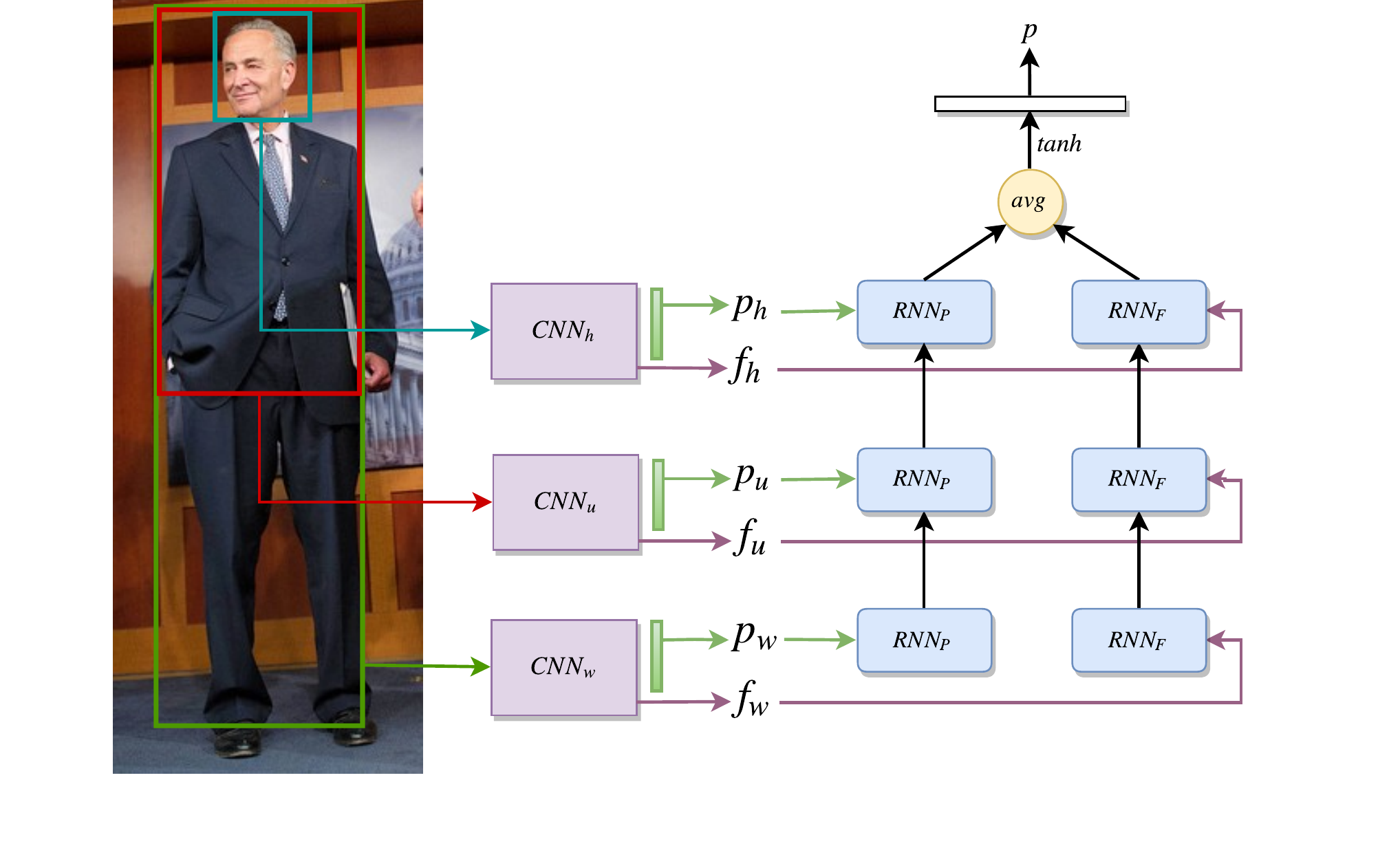}
\caption{Illustration of the proposed model. A convolutional neural network is trained for each body region. Outputs of three networks are fused using a recurrent neural network, where feature and probability vectors are processed by distinct RNN branches.}
\label{fig:model}
\end{figure}

In this paper, we tend to recognize persons in a given photo. The input to our model is an image and bounding boxes for the heads of persons in the image. As output, a label will be assigned to each person in the given input. Our general model is depicted in Figure.~\ref{fig:model}. Given an image and $B_h$ as the bounding box of the head region, bounding boxes for the upper body $(B_u)$ and the whole body $(B_w)$ are extracted. Having $B_h$, $B_u$, and $B_w$, three CNNs noted as $CNN_h$, $CNN_u$ and $CNN_w$ previously trained to identify a person based on the head, upper body and whole body regions, respectively, are applied to the corresponding extracted regions. The outputs of each CNN are a probability vector assigning probabilities to all possible identities and a feature vector giving a representation of the given region. The feature and probability vectors generated from CNNs are given to two distinct RNN branches. At the next step, the outputs of RNNs are aggregated through an averaging gate. The averaged vector is sent to a final layer after applying an element-wise $tanh$ function. The final outcome is a vector giving the probability of each identity. More detailed description of our approach is given in the following.

To train the CNN components of the model on an input image with the bounding box of the head $B_h$, bounding boxes for upper body ($B_u$) and whole body ($B_w$) of the person are extracted in a similar approach to \cite{kumar2017pose}.

Formally, if the size and location of $B_h$ are $(w,h)$ and $(l_x,l_y)$, respectively, the size and the location of $B_u$ are $(2\alpha, 4\alpha)$ and $(l_x - 0.5\alpha, l_y)$, where $\alpha = min(w,h)$. For $B_w$, the location is the same as $B_u$, but the size is $(2\alpha, 7\alpha)$.

After extracting bounding boxes for all three body parts, each CNN is trained with the corresponding image region as the input and the human identity as the output.
The CNNs are trained using the multi-class cross entropy loss defined as:
\begin{align}
  \mathcal{L} & = -\frac{1}{N_C}\sum_{i=1}^{N_C} y_i log\left(p_i\right) ,\label{eq:cross_ent}
\end{align}

where $y$ is the one-hot-encoded ground truth label for the input image, $p$ represents the softmax output vector produced by the CNN, and $N_C$ is the number of possible classes (identities).
Next, we have ${CNN}_h$, ${CNN}_u$, and ${CNN}_w$ trained on head, upper body, and whole body, respectively. For each sample, feature vectors $f_h$, $f_u$, and $f_w$ are extracted from the last layers before classification layers of the trained CNNs. We also extract the $N_C$-dimensional probability vector whose $n$-th element indicates the probability that the instance belongs to the $n$-th identity. Probability vectors for each region are extracted and noted as $p_h$, $p_u$, and $p_w$.

To combine information obtained from different body parts, we propose using RNNs in a zooming in fashion from the whole body to the upper body and then to the head to generate more confident predictions. Two distinct RNNs with equal output dimensions are used for the feature vector ($RNN_f$) and the probability vector ($RNN_p$). $RNN_f$ takes $f_h$, $f_u$ and $f_w$ as its input and likewise, $RNN_p$ receives $p_h$, $p_u$ and $p_w$ as input. 

We choose Gated Recurrent Unit (GRU) \cite{chung2014empirical} as our recurrent network architecture for its high capability of learning sequential data. Assuming $x_t$ as the input for a GRU cell at time $t$, the cell activation can be formulated as below:
\begin{align}
  {r}_{t}^{j} & = \sigma{\left({W}_{r}{x}_{t}+{U}_{r}{h}_{t-1} \right)}^{j} ,\label{eq:reset_gate}\; \\
  {z}_{t}^{j} & = \sigma {\left({W}_{z}{x}_{t}+{U}_{z}{h}_{t-1}\right)}^{j} ,\label{eq:update_gate}\; \\
  {\Tilde{h}}_{t}^{j} & = {tanh}{\left({W}{x}_{t}+{U}\left({r}_{t}\odot{h}_{t-1}\right)\right)}^{j} ,\label{eq:candidate_h}\; \\
  {h}_{t}^{j} & = \left(1-{z}_{t}^{j}\right){h}_{t-1}^{j}+{z}_{t}^{j}{\Tilde{h}_{t}^{j}},\label{eq:hidden}
\end{align}
where $\sigma$ stands for sigmoid function, $W$ and $U$ are weight matrices, and $\odot$ used in \ref{eq:candidate_h} is element-wise multiplication. ${z}_{t}^{j}$ and ${r}_{t}^{j}$ are update and reset gates at time $t$. ${h}_{t}^{j}$ and $\Tilde{h}_{t}^{j}$ calculated in \ref{eq:candidate_h} and \ref{eq:hidden} are hidden and candidate hidden vectors at time t. The value of reset and update gates are computed according to \ref{eq:reset_gate} and \ref{eq:update_gate}. The role of the reset gate is to control combination of new input and former memory. Similarly, update gate controls the amount of previous memory to keep. The value of ${h}_{t}^{j}$ will be updated using former and candidate hidden values.


With the features and probabilities as input to each RNN, final outputs of RNNs are combined as follows:
\begin{align}
  o & = tanh(average(h_p, h_f)) ,\label{eq:output}
\end{align}
where $h_p$ and $h_f$ are the outputs of the probability and feature RNNs, respectively. A final layer is added for classification. The output of the classification layer is a vector named $p$ with the size equal to the number of classes. we apply the cross entropy loss (Eqn. \ref{eq:cross_ent}) to train our model.

\section{Experiments}\label{experiments}
In this section, first, we present information about the dataset used for evaluation alongside with the specific implementation details of our approach. Then, we will provide the results of our experiments and compare the performance of our model with those of the baseline and the state-of-the-art methods.

\subsection{Dataset Description}\label{dataset_desc}
We conduct our experiments on People In Photo Album (PIPA) \cite{zhang2015beyond} dataset. PIPA contains public photo albums from users on Flickr, with their head region annotated. Head bounding boxes may be partially or fully outside of the image. It is also decided in PIPA protocol to tag no more than 10 people in a single image, meaning that not everybody in images of crowds is tagged.

Original split of the dataset consists of three parts, \textit{train}, \textit{validation} and \textit{test}. For each identity, samples are roughly partitioned in 50-25-25 percentage for the three parts respectively, with the test set consisting of 7868 images. We will use train set only to learn representations for regions of interests as described in Section \ref{approach}.
As proposed in \cite{zhang2015beyond} and followed in previous studies \cite{oh2015person,li2016multi,li2017sequential,kumar2017pose,liu2017rethinking} on PIPA, test set has been randomly split in half to \textit{$test_0$} and \textit{$test_1$} and we will follow this protocol. As mentioned in \cite{li2016multi}, there are some mislabeled instances in the test set, but to keep our results comparable with the existing methods, we won't refine the original split.

Due to the limitation of \textit{original} split proposed by \cite{zhang2015beyond}, three more challenging splits were introduced in \cite{oh2015person}, namely \textit{album}, \textit{time} and \textit{day}. In the album part, samples are collected from different albums of a person, meaning that \textit{$test_0$} and \textit{$test_1$} are sampled from different events and occasions. Time split aims to emphasize the temporal dimension of \textit{$test_0$} and \textit{$test_1$}. The metadata of photos is used to partition by newest and oldest images of an identity. Finally, day split is to challenge the appearance change. This split is made manually and date changes like seasons or visible changes like hairstyle are taken into consideration. Unlike the first three splits, the number of unique identities in day split is reduced from 581 to 199 with about 20 samples per identity.

\subsection{Implementation Details}\label{implementation_details}
Inception-V3 \cite{szegedy2016rethinking} is the architecture of choice for the CNNs in our model. We initialize CNNs with the weights of the pre-trained model on the ImageNet and for each body part, CNNs are trained on the train split. This pre-training step injects additional data with a similar distribution to test split of PIPA into the CNNs and helps them perform better when trained on $test_0$ or $test_1$. In pre-training step, we train each CNN for 50 epochs using Stochastic Gradient Descent (SGD) optimizer with a learning rate of 0.01 and momentum of 0.9. To train CNNs on each half of the test split of PIPA, we initialize networks with pre-trained weights obtained by training on train split. Here, CNNs are trained for 50 epochs with a learning rate of 0.01 and 20 epochs with learning rate of 0.001. Again, we use SGD with a momentum of 0.9. 
All input images are resized to the fixed size of 299$\times$299. During training, we use various methods to augment the dataset. Images are randomly flipped. Random rotation with the range of 30 degrees, is done. Horizontal and vertical shifts of -60 to 60 pixels are performed randomly. Zooming in or out is also performed in the range of 0.8 and 1.2 of the image size.

We use a GRU with three timesteps and 2048 output dimensionality. Drop out with a probability of 0.5 is applied to the combined representation of GRUs. SGD with a learning rate of 0.005 and momentum of 0.9 is used to optimize the loss. The number of training epochs is fixed on 2000. Training this part for each fold of test split takes about 1.5 hours on a single Geforce GTX 1080 Ti NVIDIA GPU. We use Keras \cite{chollet2015keras} with Tensorflow \cite{abadi2016tensorflow} backend for our implementations. 

\subsection{Experimental Results}\label{exp_results}
Now we explore the importance of modeling the relational cues of different body regions and good practices in usage of recurrent architectures for this purpose. All reported results throughout the paper are classification accuracies averaged over \textit{$test_0$} and \textit{$test_1$}, meaning that each model has been trained on \textit{$test_0$} and evaluated on \textit{$test_1$} and vice versa.

\textbf{Initial CNN predictions.} In the first stage of our work, we train CNNs on each body part. Every part-specific CNN can classify a person on its own. In Table.~\ref{baselines}, accuracies of predictions from body part specific CNNs along with their average and maximum fusion variants are summarized. It is evident that an increase in body part size makes it harder for the model to perform well and as expected, the most informative single region is the head of the person. When we fuse predictions of different body parts with either element-wise average or maximum, the accuracy of the model increases noticeably in all splits except the day split. This validates the idea that different cues are present in each body part, which can be extracted by fusion methods. In the day split, performances of the upper and the whole body CNNs have a large gap with the head CNN which makes simple fusion methods yield poor predictions.

\begin{table}[]
\begin{tabular}{@{}ccccc@{}}
            & Original   & Album  & Time   & Day \\ \midrule
Whole Body   & 81.73  & 71.28  & 59.15  & 31.13 \\
Upper Body   & 85.36  & 76.07  & 64.49  & 36.40 \\ 
Head         & 86.40  & 80.29  & 70.90  & \underline{\textit{54.98}} \\
Element-wise Avg  & \underline{\textit{89.68}}  & 82.37  & 72.81  & 50.95 \\
Element-wise Max      & 89.57  & \underline{\textit{82.52}}  & \underline{\textit{73.13}}  & 53.23 \\ \midrule
Ours         & \textbf{90.88}  & \textbf{84.40}  & \textbf{76.44}  & \textbf{56.92} \\
\end{tabular}
\vspace{0.1cm}
\centering
\caption{Baseline performance comparison. Evaluations of each part-specific CNN reported and compared to simple fusion methods in order to provide a simple baseline for our task. Classification accuracies (\%) are reported and the top two results of each method are marked in \textbf{bold} and \underline{\textit{italic}}}
\label{baselines}

\end{table}

\begin{figure*}
\centering
\includegraphics[trim = 40 670 40 0,clip,height = 5cm]{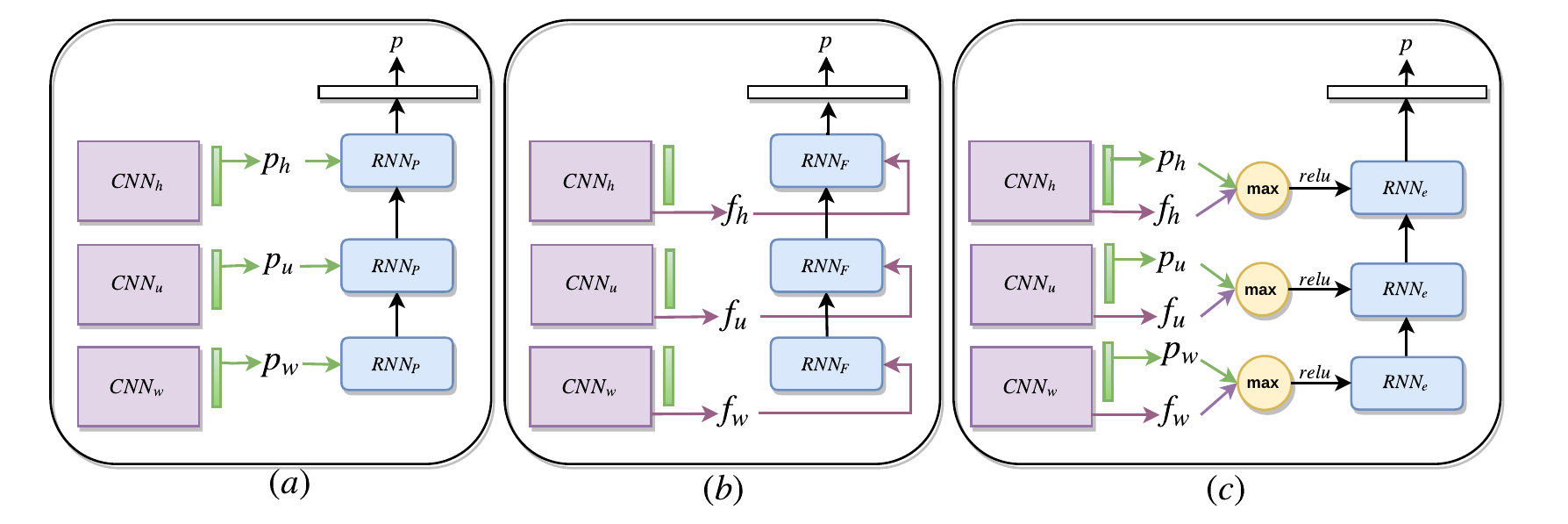}
\caption{Different variations of recurrent architecture tested for CNN output fusion. (a), (b) and (c), are the illustrations of Probabilities RNN, Features RNN and Embeddings RNN, respectively}
\label{fig:model_variations}
\end{figure*}

\textbf{Fusion Baselines.} Here, we analyze more complex baselines to combine information from different body parts. We experiment with different versions of our RNN-based model as shown in Fig.~\ref{fig:model_variations}.

\begin{itemize}
  \item \textbf{Concat: } Concatenated features from all CNNs are fed into a fully connected layer with 2048 neurons and a classification layer on top of it.
  \item \textbf{Confidence-Aware: } Similar to \cite{li2016multi}, a weighted average of probabilities of different body parts with respect to the confidence of predictions is calculated as final output.
  \item \textbf{Probabilities RNN: } A variant of our model where only one RNN is used on input probabilities produced by CNNs to produce final predictions.
  \item \textbf{Features RNN: } Similar to probabilities RNN but with CNN features as the input.
  \item \textbf{Embeddings RNN: } Features and probabilities of each CNN are combined in an embedding layer. Outputs of the embedding layer are given to a single RNN to identify persons. Embedding layer has a fully connected layer on top of probabilities and features to embed them to a new fixed-size layer. Outputs of the fully connected layers are combined using element-wise maximum and a $relu$ activation function on top of it to form the output of the embedding layer. In this case, we observed that this way of combining embeddings works better than other methods like average and $tanh$ activation function.
  \item \textbf{Reversed Zoom-RNN: } Similar to our final approach, except that the head region is the input to the first timestep of the RNN and the whole body is the last.
  \item \textbf{Zoom-RNN: } The complete version of our model.
\end{itemize}

Results of the baselines are reported in Table.~\ref{fusion_methods}. \textit{Concat} is able to combine information of body parts to some extent but it is not able to perform better than previous simple fusion methods. \textit{Confidence-Aware} gives the best aggregation result in \cite{li2016multi}, but all of our RNN variations outperform it in all of the splits. \textit{Probabilities RNN} reasons over prediction probabilities from least confident to most confident. Like the previous baseline, although it has the ability to fuse some information from three predictions, it shows worse performance than simple average or element-wise maximum. Letting the model learn from visual features in \textit{Features RNN} increases the performance over simple fusion methods. Furthermore, to evaluate whether using both probability and feature vectors is beneficial or not, the accuracy of the \textit{Embeddings RNN} is reported. In this way, performance is slightly worse than \textit{Features RNN} in most of the splits. We believe combining probabilities and feature vectors in lower level representations is not able to produce a strong combination. In \textit{Zoom-RNN}, features and probabilities of CNNs are separately encoded into higher level representations and a combination of these representations is made. Significant improvement of accuracy in all four splits over other baselines and previous fusion methods proves our statement about the combination of probabilities and features in higher levels of representation. 

Here, another important factor in the combination using RNN is the order of input CNNs. The poor performance of \textit{Reversed Zoom-RNN} indicates that starting from the best performing part-specific CNN to worst one can make it difficult for the model to make true inferences. Although because of the improvement over the worst part, it is obvious that the model remembers some information about other parts, but it is also evident that most of the valuable cues are forgotten. Therefore a good practice is to start from weakest part-specific model to strongest one to make it easier for the recurrent model not to forget the best performing model's representations and also remember some valuable information from other body parts. 

\begin{table}[]
\centering
\begin{tabular}{@{}ccccc@{}}
            &Original&Album&Time&Day \\ \midrule
Concat          & 88.62  & 80.16  & 70.03  & 47.10 \\
Confidence-Aware\cite{li2016multi}          & 
89.56  & 82.19  & 72.55  & 50.11 \\
Probabilities RNN   & 88.44  & 80.67  & 71.05  & 52.41 \\
Features RNN      & \underline{\textit{89.68}}  & \underline{\textit{83.56}}  & \underline{\textit{75.25}}  & 55.19 \\
Embeddings-RNN          & 89.32  & 83.23  & 74.82  & \underline{\textit{55.57}} \\
Reversed Zoom-RNN         & 86.54  & 77.93  & 66.08  & 37.34 \\\midrule
Zoom-RNN (Ours)         & \textbf{90.88}  & \textbf{84.40}  & \textbf{76.44}  & \textbf{56.92}\\
\end{tabular}
\vspace{0.1cm}
\caption{Evaluation of different fusion methods and proposed method. Results indicate that our final method outperforms all the variations}
\label{fusion_methods}
\end{table}

\subsection{Comparison to The State-of-the-Art}

As discussed in Section \ref{related}, there have been various approaches in person recognition on PIPA. The results are summarized in Table \ref{stateofthearts}. 
Our model has better performance on all four splits of PIPA compared to PIPER \cite{zhang2015beyond}, Sequential \cite{li2017sequential}, naeil \cite{oh2015person}, and Pose-Aware \cite{kumar2017pose}. It also outperforms \cite{li2016multi} when they don't use additional contextual cues but it can't perform better than their model using contextual cues in the day split.

We are aware that a recent study \cite{liu2017rethinking} performs better in three splits out of four, by using a novel loss function (COCO). Given that it uses an additional face region, to be able to compare the results on the same body regions, we report \cite{liu2017rethinking} with 3 body regions, which is outperformed in all four splits of PIPA by our method. However, in this work, we are interested in showing that our relational modeling of body regions in a zooming fashion from the broader region to the detailed and more informative one improves baseline performances and it is not necessarily in conflict with other approaches like \cite{liu2017rethinking}.

Unlike \cite{li2016multi}, our main model does not use any contextual information other than body parts of the person, so we expect a better performance by taking advantage of additional cues. Therefore, we have implemented our version of inter-person sequence similar to \cite{li2017sequential}. The positive effect of inter-person sequence as means of adding contextual information of co-occurrence of the persons shows that further improvements on our model are possible.
\begin{table}[]
\label{stateofthearts}
\scalebox{0.9}{
\begin{tabular}{@{}ccccc@{}}
& Original   & Album  & Time   & Day \\ \midrule
PIPER\cite{zhang2015beyond}               & 83.05   & -   & -   & -  \\
Sequential\cite{li2017sequential}          & 84.93   & 78.25   & 66.43   & 43.73  \\ 
naeil\cite{oh2015person}               & 86.78   & 78.72   & 69.29   & 46.61  \\
\cite{li2016multi} w/o context       & 83.86   & 78.23   & 70.29   & 56.40  \\
\cite{li2016multi} with context      & 88.75   & 83.33   & 77.00   & \underline{\textit{59.35}}  \\
Pose-Aware\cite{kumar2017pose}          & 89.05   & 82.37   & 74.84   & 56.73  \\\midrule
COCO\cite{liu2017rethinking}         & \textbf{92.78}   & {83.53}   & \textbf{77.68}   & \textbf{61.73}  \\
COCO\cite{liu2017rethinking} with 3 body regions    & 89.71   & 78.29   & 66.60   & 52.21  \\\midrule
Ours         & {90.88}  & \underline{\textit{84.40}}  & {76.44}  & 56.92 \\
Ours + inter-person sequence         & \underline{\textit{91.36}}  & \textbf{85.00}  & \underline{\textit{77.11}}  & 58.53 \\ 
\end{tabular}}
\vspace{0.1cm}
\caption{Comparison with state-of-the-arts.}
\end{table}

\vspace{0.5in}
\section{Visualization}\label{visualization}
To illustrate our model's zooming nature and the effect of modeling relational cues of different body regions, in this section, we provide examples of our predictions on PIPA test set.

\begin{figure}
\centering
\includegraphics[trim = 30 350 0 5,clip,height = 5cm]{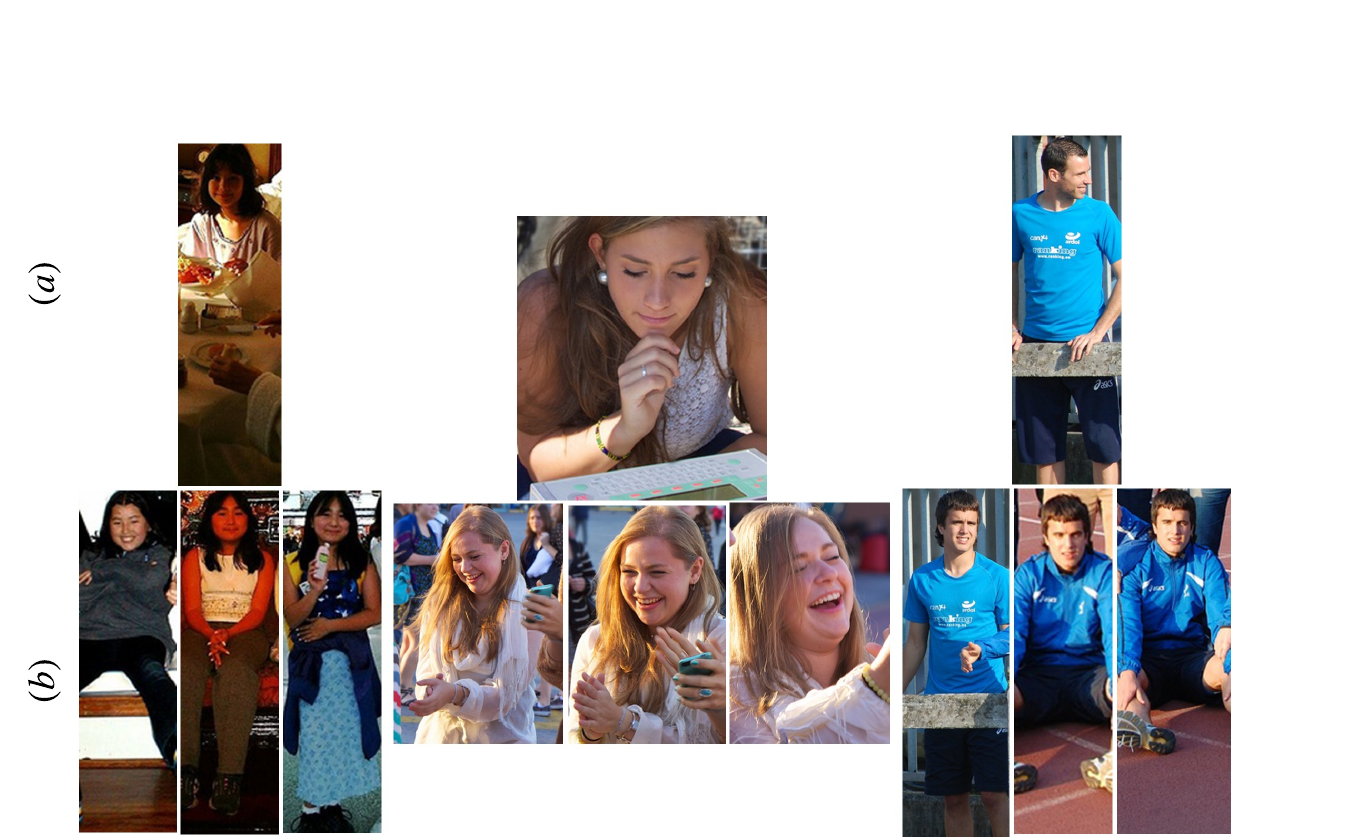}
\caption{Examples where average fusion method confuses the identity while our model predicts accurately. (a), is the input image and three images of the confused identity have been shown below each input image in row (b)}
\label{fig:average_wrong}
\end{figure}

In Figure \ref{fig:average_wrong}, we show examples that average method mislabels the identity, while our model predicts the right one. It can be inferred that similar outfit and faces can easily misguide the averaging methods, while taking advantage of relation of the body regions using our model performs accurately.

\begin{figure}
\centering
\includegraphics[trim = 38 350 0 30,clip,height = 5cm]{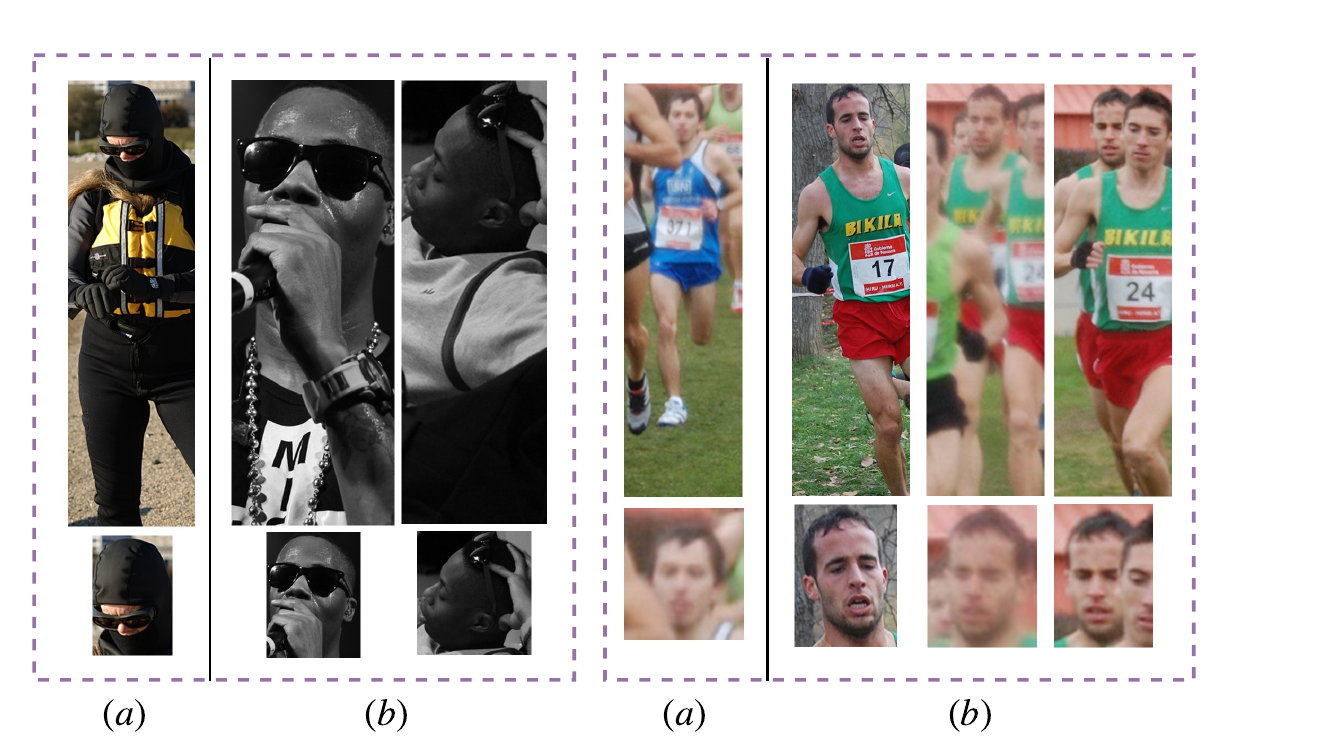}
\caption{Examples where the predictions have been wrong using only head features, while taking advantage of different cues is effective. The first column (a) of each example is the input image and the next ones (b) are the confused identities. The head regions of each image are shown below the image.}
\label{fig:head_wrong}
\end{figure}

Similarly, in Figure \ref{fig:head_wrong}, we show some instances in which head features alone may misguide the model, but taking advantage of the information from different body parts helps our model predict accurately. As mentioned in Section \ref{introduction}, person recognition task in photo album includes challenges like non-frontal face, occlusion and motion blur. It can be understood from the examples in Figure \ref{fig:head_wrong} that we can overcome these challenges by extracting good information from different body regions.

\section{Conclusions}\label{conclusion}
In this paper, we proposed a novel method for combining cues of different body regions for the task of person recognition in photo album. Our approach uses two distinct recurrent neural networks to extract information present in different parts of a human photo in order to improve recognition performance. We conduct experiments on PIPA dataset and show that our model significantly boosts baseline performances. We also achieved state-of-the-art results in one split and second-best results in others by a narrow margin while not using contextual cues which have been proved to significantly increase the overall performance.

{\small
\bibliographystyle{ieee}
\bibliography{egbib}
}

\end{document}